\documentclass[conference]{IEEEtran}
\usepackage{times}
\usepackage{graphics,amssymb,amsmath,epsfig,color}
\usepackage{ltxgrid}
\usepackage{graphicx}
\usepackage{textcomp}
\usepackage{enumitem}
\usepackage{balance} 
\usepackage{xcolor}
\usepackage[numbers]{natbib}
\usepackage[bookmarks=true]{hyperref}

\usepackage{comment}
\usepackage{hhline}
\usepackage{caption}
\usepackage{subcaption}
\usepackage{amsmath}
\usepackage{tikz}
\usetikzlibrary{positioning}
\usepackage{pgfplots}
\usepackage{pgfplotstable}
\pgfplotsset{compat=1.7}
\usepgfplotslibrary{groupplots}
\usepackage{multirow}

\usepackage{comment}

\def\BibTeX{{\rm B\kern-.05em{\sc i\kern-.025em b}\kern-.08em
    T\kern-.1667em\lower.7ex\hbox{E}\kern-.125emX}}

\makeatletter
\newenvironment{customlegend}[1][]{%
    \begingroup
    \pgfplots@init@cleared@structures
    \pgfplotsset{#1}%
}{%
    \pgfplots@createlegend
    \endgroup
}%

\def\addlegendimage{\pgfplots@addlegendimage}
\makeatother

\begin{document}

\title{Robot Duck Debugging: Can Attentive Listening Improve Problem Solving?}

\author{Author Names Omitted for Anonymous Review. Paper-ID [add your ID here]}

\author{
\authorblockN{Maria Teresa Parreira}
\authorblockA{KTH Royal Institute of Technology\\
Stockholm, Sweden}
\and
\authorblockN{Sarah Gillet}
\authorblockA{KTH Royal Institute of Technology\\
Stockholm, Sweden}
\and
\authorblockN{Iolanda Leite}
\authorblockA{KTH Royal Institute of Technology\\
Stockholm, Sweden}
}

\maketitle
\twocolumngrid
\begin{abstract}
While thinking aloud has been reported to positively affect problem-solving, the effects of the presence of an embodied entity (e.g., a social robot) to whom words can be directed remain mostly unexplored.
In this work, we investigated the role of a robot in a "rubber duck debugging" setting, by analyzing how a robot's listening behaviors could support a thinking-aloud problem-solving session. Participants completed two different tasks while speaking their thoughts aloud to either a robot or an inanimate object (a giant rubber duck). We implemented and tested two types of listener behavior in the robot: a rule-based heuristic and a deep-learning-based model. In a between-subject user study with 101 participants, we evaluated how the presence of a robot affected users' engagement in thinking aloud, behavior during the task, and self-reported user experience. In addition, we explored the impact of the two robot listening behaviors on those measures. In contrast to prior work, our results indicate that neither the rule-based heuristic nor the deep learning robot conditions improved performance or perception of the task, compared to an inanimate object. We discuss potential explanations and shed light on the feasibility of designing social robots as assistive tools in thinking-aloud problem-solving tasks.

\end{abstract}

\IEEEpeerreviewmaketitle


\section{Introduction}


The concept of "Rubber Duck Debugging" originated from a story in the book \textit{The Pragmatic Programmer} \cite{rubberduckbook}. It encompasses the idea that a rubber duck can assist the process of "debugging" a program by serving as an embodied listener when a person explains the code aloud step-by-step. This concept is related to the technique of Think-Aloud Problem Solving (TAPS) \cite{2010Ku}, which consists of the verbalization of one's thoughts while solving tasks. Prior literature found that verbalization is an effective strategy to improve learning and task performance \cite{1985stinessen,1994michelene}, but this activity takes place as an individual process. Rubber duck debugging reframes the act of problem-solving as a thinking-aloud interaction, where the second interactant is an unresponsive listener. 

Social robots can assist people in therapy for rehabilitation \cite{butchart2021child, casas2021social} and mental health \cite{rabbitt2015integrating, kabacinska2021socially}, or support children with special needs \cite{scassellati2012robots, dickstein2018socially, martinez2020socially}. More specifically, the presence of a robot in a TAPS session was seen to improve learning gains in young students \cite{2018ramachandran}. Moreover, work by \citet{Gratch2006} showed that a responsive listener agent, exhibiting non-verbal behavior, leads subjects to spend more time talking and use more words, when compared to an unresponsive listener agent. 

In this work, we explored a "rubber duck problem-solving session" and the impact of replacing the inanimate object with an active listener robot. 
Prior work has established different methods to generate listener behaviors, formulating rule-based heuristics extracted from human conversation data \cite{2010truong,2021oertel} or using deep learning (DL) models \cite{2019huang,2019wijnen}. We aim to examine if listener behaviors generated through either a heuristic or a DL model lead to different outcomes in a TAPS session. This allows us to explore the sensitivity of the user to robot behavior when the focus of the session is on solving the task rather than the social interaction with the robot.

\begin{figure}
    \centering
    \includegraphics[width=0.37\textwidth]{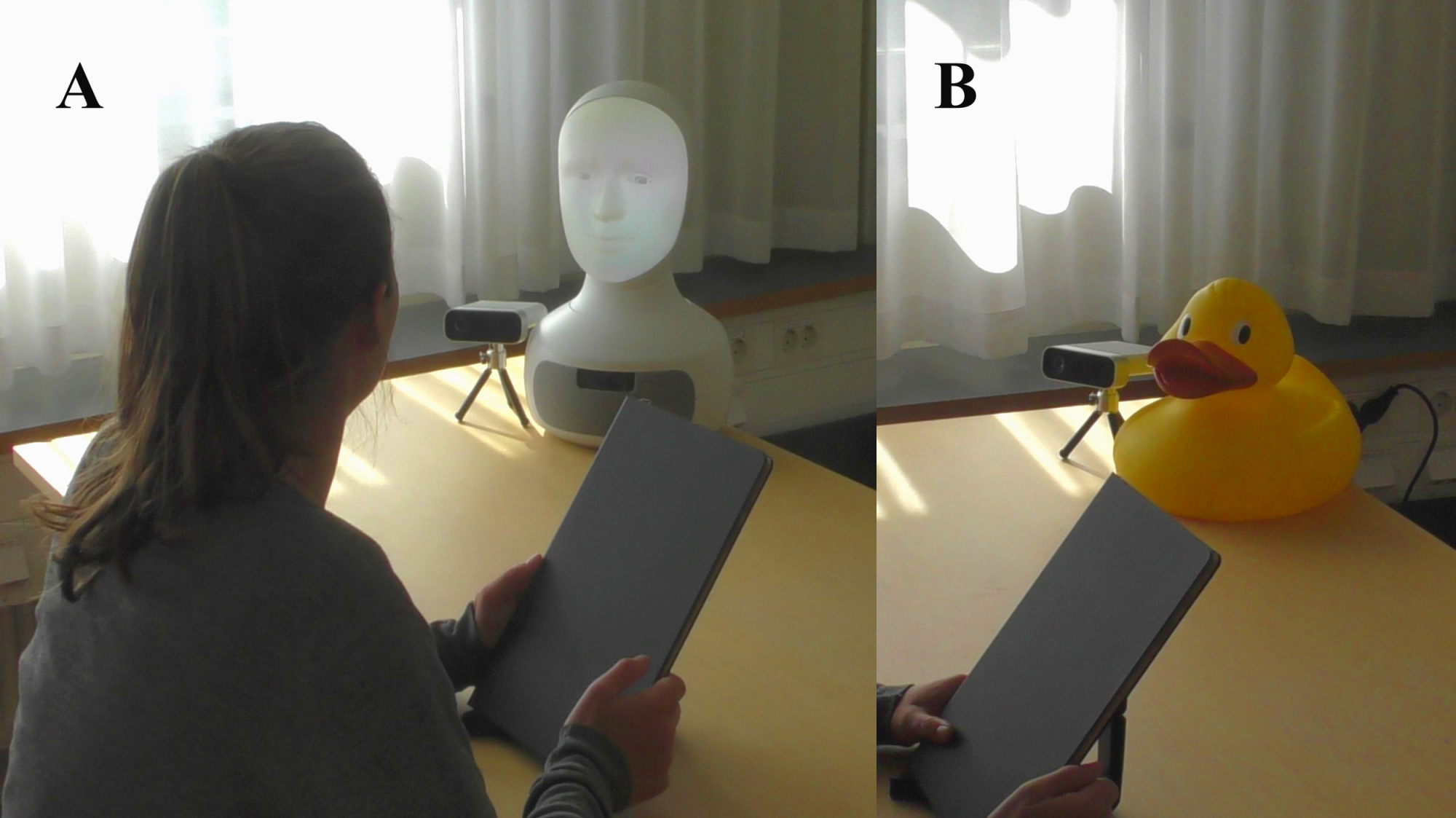}
        \caption{Overview of the experimental set-up. Participants were invited to think aloud while completing tasks on the tablet. They directed their words to A) a Furhat robot which displays either a heuristic (\textit{NaïveL}) or learned (\textit{DataL}) listening behavior or B) an inanimate object (giant rubber duck).
    }
    \label{fig:setup}
    \vspace{-1em}
\end{figure}

In a between-subject study with three conditions, we evaluated how the two different robot listening behaviors and an inanimate object - a rubber duck - affect the outcome and perception of two distinct think-aloud problem-solving tasks, a deductive logic quiz and an open-ended question. To the best of our knowledge, this is the first study characterizing how the presence of an active listener robot affects the perception of a TAPS session. Considering the reported positive effects of verbalization during problem-solving and the presence of robots in learning scenarios, we explore the potential of robots as "social assistants" in everyday problem-solving tasks. In sum, our main contributions are: (1) investigating the impact of an active listener robot in a thinking-aloud task; (2) developing and evaluating a rule-based listening behavior and a deep-learning based listening behavior system from publicly available human-human dyadic conversational data; (3) exploring how methods and datasets used for developing listener behaviors in one social setting - a conversation - perform in the context of attentive listening to assist a problem-solving session.

\section{Related Work}
\label{sec:relwork}

This research is based on previous work in the areas of social robots in think-aloud tasks and generating active listening behavior.
\begin{description}[align=left, leftmargin=0em, labelsep=0.2em, font=\textbf, itemsep=0em,parsep=0.3em]
\item[Social Robots in Thinking-aloud tasks:] Prior literature has explored how a robot can play an auxiliary role in problem-solving. \citet{2019wijnen} studied how the presence of a social robot can affect explanatory behavior in children, which is associated with deeper learning. The authors concluded that the presence of a robot to which children could direct their words was effective in triggering explanatory behavior (longer duration of utterances, more relevant content). Closer to our work, \citet{2018ramachandran} explored how the presence of a social robot while engaging in a TAPS task could improve learning gains in young students, versus a no-robot condition. The robot would intervene with tutoring notes and reminders to keep thinking aloud. They concluded that the robot platform could support the thinking-aloud process and enhance immediate and persistent learning gains. The authors suggest that the robot represented an embodied social entity towards which students could direct their words, which is in line with the findings from \citet{2019wijnen}. To the best of our knowledge, there is no similar work on adult subjects.

 We expand on prior work by comparing the effects of the presence of an active listener robot to those of an inanimate object in a "rubber duck debugging"/TAPS session with adults. In order to assess if different robot listening behaviors lead to different outcomes, we implemented two types of active listener behaviors, both based on previous work.

\item[Generating Listening Behavior:]

Listening behavior is characterized by backchannels (BCs) \cite{1970yngve}, which signal attentiveness or emotion to a speaker. These include body language, such as head nods or smiles, and short vocalizations (\textit{'uh-huh', 'hmm'}). Backchannel placement is an important component of conversations, as it can skew the perception of the interactants~\cite{2010poppe}.

Significant amount of work has studied the generation of listening behavior in robots.
One approach is to develop rule-based methods that are informed by prosodic features, such as voice pitch and pauses \cite{2010truong,2000ward}, or by behavior observed in human-human conversational data \cite{2021oertel}. As an alternative to designing hand-crafted heuristics, the use of automated methods has been widely adopted\cite{2013dekok, 2016kawahara}. \citet{1996okato} designed a Hidden Markov Model (HMM) based on prosodic patterns to detect when to emit a BC. \citet{2010morency} also used an HMM with multimodal (audiovisual) input features to predict BC timing. 

 More recently, deep learning techniques have been deployed to inform the design of listening behaviors. Examples include Recurrent Neural Network (RNN) models, which capture the temporal dependencies of continuous signals (i.e., retain "memory" of previous inputs). \citet{2019ruede} used Long-Short Term Memory (LSTM) layers to predict BCs, with acoustic features and word history as input features. \citet{2019huang} also tested LSTMs and Gated Recurrent Unit (GRU) layers for the development of idle behavior in virtual agents, with multimodal input features. Similarly, \citet{2021murray} made use of LSTMs and multimodal input, and suggested a method for data augmentation that positively impacts BC prediction. 
 While the use of RNNs has been widely adopted in sequential decision-making tasks, such as the generation of listener behavior, we highlight other interesting approaches, such as semi-supervised learning \cite{jain2021exploring}, or reinforcement learning \cite{2019hussain,2019rudovic}.

In the present work, we developed two different robot listener behavior generation policies - one rule-based, one DL-based - and evaluated how these impacted various aspects of think-aloud problem-solving tasks: engagement with thinking-aloud, task performance, and user experience. 

\end{description}

\section{Implementation of Listener Behavior} \label{sec:listener}

We set out to study how different listening behaviors can support a "rubber duck" problem-solving session. The listening behavior consisted of the robot's \textbf{gaze} and \textbf{backchanneling} (vocal or non-vocal). For \textit{vocal} utterances, the specific sound was selected randomly from a set of pre-defined utterances (\textit{e.g., 'hmm', 'ahh'}). The \textit{non-vocal} backchannel was realized through a head nod. 

For developing backchanneling behavior models, we included two different approaches, as we intended to shed light the sensitivity of the user to listening behaviors when the interaction with the robot is only \textit{auxiliary} (not \textit{collaborative}) to task completion. 
We tested each of the two models as a different robot listening condition in our user study. 

As suggested by prior work~\cite{jain2021exploring}, the decision-making process for generating a backchannel was divided into two stages: first, the \textit{timing} of the backchannel was determined; in a second step, the \textit{type} of backchannel was selected. This separation allows for handling the natural imbalance in conversational data (few examples of each backchannel type and more datapoints for "no backchannel").

Backchannel timing and type were either heuristically determined (naïve listener model, \textit{NaiveL}) or learned through the application of a deep learning model (data-driven model, \textit{DataL}). We detail the implementation of these behaviors below.

\subsection{Naïve Listener}
\label{sec:naivel}

To develop the robot's behavior for the naïve listener condition, we decided to use a frequently deployed heuristic, originally developed by \citet{2000ward}. This work used a corpus of human conversational data in English to implement a heuristic that finds the most appropriate \textit{timing} for a backchannel in a dyadic interaction. The heuristic is based on prosodic cues, i.e. pitch. The authors define a set of conditions for producing a backchannel, which we reproduce \textit{verbatim}:
\textit{\begin{itemize}
    \item[\textbf{(P1)}] a region of pitch less than the 26th-percentile pitch level and
    \item [\textbf{(P2)}] continuing for at least 110 milliseconds,
    \item [\textbf{(P3)}] coming after at least 700 milliseconds of speech,
    \item [\textbf{(P4)}] providing you have not output backchannel feedback within the preceding 800 milliseconds,
    \item [\textbf{(P5)}] after 700 milliseconds wait.
\end{itemize}
}

To run this model in our user study, we computed the pitch distribution of the user's voice over the last $50$ s of audio and calculated the pitch region percentile over the last $10$ ms. 
The backchannel \textit{type} was randomly selected to be either vocal or non-vocal when the five conditions shown above were met.

\subsection{Data-Driven Listener}
\label{sec:datal}
We explored the use of machine learning to determine backchannel timing and type. While we maintain the input modality in the form of audio features, as in the naïve condition, we wanted to explore if the learned behavior, by having higher potential to adapt to different speakers, could lead to more positively evaluated interactions \cite{2010morency}. 

\begin{table*}[]
\centering
\tabcolsep=0.11cm
\caption{Performance metrics on model used (averaged over all validation folds). Both models used augmented data for training. Comparison with performance of the rule-based model (\textit{NaïveL} condition) on the same dataset.}
\label{tab:modelperf}
\begin{tabular}{c|c|c|c}
\hline
\textbf{Module}                & \textbf{Model}                & \textbf{Hyperparameters}                                                                                                                                                                       & \textbf{Performance}                                         \\ \hline
\multirow{3}{*}{\textbf{Timing}} & \multirow{3}{*}{\textit{GRU}} & \multirow{3}{*}{\textit{\begin{tabular}[c]{@{}c@{}}Lookback: 5,activation: sigmoid, \\ batch size: 16,  dropout: 0.0, \\ loss function: focal,optimizer: Adam\end{tabular}}} & \textbf{Macro Accuracy}: 0.95, \textbf{Precision}: 0.52, \textbf{Recall}: 0.51, \textbf{F1}:0.50  \\
   &    &   & \textbf{Margin Accuracy}: 0.95, \textbf{Precision}: 0.59, \textbf{Recall}: 0.76,  \textbf{F1}:0.65 \\
                               &                               &                                                                                                                                                                                       & \textbf{BC Prediction Deviation}: 0.83                                       \\ \hline
\multirow{3}{*}{\textbf{Type}} & \multirow{3}{*}{\textit{GRU}} & \multirow{3}{*}{\textit{\begin{tabular}[c]{@{}c@{}}Lookback: 10, activation: sigmoid, \\ batch size: 32, dropout: 0.2,\\  loss function: MSE, optimizer: SGD\end{tabular}}}   & \textbf{Macro Accuracy}: 0.64, \textbf{Precision}: 0.37, \\    &    &         &       \textbf{Recall}: 0.39, \textbf{F1}:0.35                                                         \\ 
                               &                               &                                                                                                                                                                                       &      \\  \hline \hline                                      
\multirow{2}{*}{\textbf{Timing}} & \multirow{2}{*}{\textit{NaïveL}} & \multirow{2}{*}{\citet{2000ward}}   & \textbf{Macro Accuracy}: 0.95, \textbf{Precision}: 0.48, \\    &    &         &       \textbf{Recall}: 0.50, \textbf{F1}:0.49 
\end{tabular}
\end{table*}

\subsubsection{Problem Formulation}
We formulated the problem of generating backchanneling in a TAPS session as a sequential decision-making problem. At any time-step $t$, the environment (user's voice features) is captured as state variable $s_t \in S$. The robot takes actions $a_t \in A$. It first chooses between a binary output: \textit{performing BC} or \textit{doing nothing}. If the robot chooses \textit{performing BC}, $s_t$ is again used to inform \textit{which type} of BC to perform. The potential actions $a_t$ that follow are \textit{vocal BC} (vocal utterance), \textit{non-vocal BC} (nodding), or \textit{both}. 

\begin{description}[align=left, leftmargin=0em, labelsep=0.2em, font=\textbf, itemsep=0em,parsep=0.3em]
\item[Dataset:]
The collection of conversational data is frequently the first step to developing listening behaviors. In this work, we wanted to test if we could instead use data collected in a different setting from that in which we deployed our model. Some interesting corpora of dyadic interactions in English include CCDb \cite{ccdb}, IEMOCAP \cite{iemocap} and D64 \cite{D64}, among others \cite{2017corpora}. We were looking for unscripted, non-topic-bounded interactions that were publicly available. As per these criteria, the dataset used for training was Cardiff's Conversation Database (CCDb)\cite{ccdb}.

The CCDb consists of 30 dyadic conversations, each lasting around five minutes. The 30 conversations are between 16 different speakers (12 M, 4 F), with ages ranging from 25-56 years old. We used the subset of conversations that were annotated for facial expressions and utterances to train the listening behavior policy. We extracted data from the perspective of each participant, totaling 80 minutes of conversational data. 

We split the individual feature streams into moments where the participant was the \textit{speaker} or the \textit{listener}. As we are focusing on listener behavior, we extracted features only in the \textit{listener} portions of the conversation. As training data, we used the audio features from the \textit{speaker} (other participant), while extracting the annotated backchannels performed by the \textit{listener} participant as ground truth for our model. The conversations from the CCDb dataset include extensive annotations. We filtered the annotations to provide positive training instances in moments where vocal backchannels or head motion (nodding) were present.

\item[State space:]

As input features for our model, we extracted speech features from the speaker, including 13-dimensional mel-frequency cepstrum coefficients (MFCC) and 4-dimensional prosody features as used in prior works \cite{2019ruede,2021murray,jain2021exploring}. The MFCC features are computed every 30 ms with a sliding hamming window of 400 ms. Prosody features include pitch (fundamental frequency) and yin-energy, as well as the first derivative of these variables. The final 34-item feature vector is composed of the mean and standard deviation of each of these features. The dataset frequency is 2 Hz (one sample every 0.5 s) and was normalized before training.

\item[Action Space:]
The robot could perform the same actions as those performed by the \textit{NaïveL} robot (Sec. \ref{sec:naivel}) - vocal utterance, nod of varying amplitude, or a simultaneous combination of both.
\end{description}

\subsubsection{Training policies for listening behavior} 
To train sufficiently good behavior policies, we explored different model architectures and techniques such as data augmentation\cite{2021murray}. Below, we explain the models tested and the selection process, as well as model performance.

\begin{description}[align=left, leftmargin=0em, labelsep=0.2em, font=\textbf, itemsep=0em,parsep=0.3em]

\item[Data Augmentation:]

\citet{2021murray} suggest a method to improve the robustness of robot listening behaviors by using audio data augmentation. The authors show that the model trained on this data outperforms a rule-based and a random model as reported by users. We emulated the proposed method by making use of \textit{masking} techniques in the time and frequency domains. In our work, training instances (audio features) were partially masked in one or both domains, chosen at random. Both the original and the deformed instances were used for training.

\item[Models:]
We explored the use of deep learning techniques to develop listening behavior. We separated our model into two components according to the two-staged decision-making process, the \textbf{Timing} and the \textbf{Type} modules. The \textbf{Timing} module learns the appropriate timing to perform backchanneling, and the \textbf{Type} module decides which type of feedback - vocal, non-vocal (nodding) or both - the robot should emit. The first module is a binary classification task (\textit{do BC} or \textit{do nothing}). In the case of a positive (\textit{'do BC'}) output, the second module is activated (a multiclass classification model).

The use of neural networks has proven to be a versatile tool to learn relevant features automatically \cite{2015mueller}. Multiple authors have approached this problem by making use of models that consider previous internal states. LSTMs are a common approach \cite{2019ruede,2021murray}, although these architectures are prone to overfitting. Alternatively, Gated Recurrent Units (GRUs) can be used \cite{2019huang}. These are less complex and usually preferred over LSTMs for small training datasets. 

\item[Model Hyperparameter Tuning:]
In order to determine the best model, we tested different combinations of models and parameters. For both the \textit{Timing} and \textit{Type} modules, we trained single-layer LSTM and GRU models, which were followed by a dropout layer and a final dense layer. Each model was either trained on the original or the augmented dataset, considering varying previous timesteps, i.e. lookback sizes (5, 10 and 15, respectively 2.5, 5 or 7.5 seconds of data), activation functions (sigmoid, ReLU and softmax), batch sizes (8, 16, 32), dropout rates (0, 0.2 and 0.4), L2 regularization (0.1, 0.01, 0.001, 0.0001), loss functions (focal loss, MSE, binary cross-entropy or hinge loss) and optimizers (SGD and Adam).

\item[Evaluation Metrics:]

Performance on both modules was evaluated based on macro \textbf{accuracy} (number of correct predictions over all predictions), \textbf{precision} (probability that emitted BCs match the ground-truth data), \textbf{recall} (probability that a ground-truth BC is predicted by the model) and \textbf{F1-score} (harmonic mean of precision and recall). 

Additionally, for the \textbf{Timing} module and in line with similar work \cite{2012dekok,2019ruede}, we calculated the above metrics considering a tolerance margin of $[-500, 500]$ ms; finally, we controlled for robot overreaction with a \textbf{Backchannel frequency deviation} metric. For each participant $i$, we considered the relative difference between the predicted number of backchannels and its original value:
\begin{equation}
\Delta{BC}^{i}_{dev} = |Y^i_{true} - Y^i_{pred}| / Y^i_{true}
\end{equation}

\item[Models' Selection and Performance:]
For training, data was split into 8 folds, with non-overlapping participant data in a 6:1:1 train-validation-test split. Each model was trained for a maximum of 100 epochs, but early stopping strategies were deployed for both loss and validation loss, in order to prevent overfitting. The use of a dropout layer and L2 regularization entailed the same purpose.

Each module (\textit{Timing} and \textit{Type}) was trained with an analogous methodology. We fine-tuned hyperparameters based on macro accuracy and F1-score. The final candidates for each model type (LSTM or GRU, augmented and not augmented training data) were then trained using k-fold cross-validation. 

For each module, we deployed the model that was most frequently in the top three best performing models for each of the metrics mentioned above. Both models are single-layer GRUs trained on the augmented dataset (see performance on Table \ref{tab:modelperf}, along with performance for the \textit{NaïveL} heuristic on the same dataset). The \textbf{Timing} model outperforms other similar model architectures for BC prediction \cite{2019ruede,2021murray}. Interestingly, the modules have different lookback periods, with the \textbf{Type} model input considering a higher number of past internal states.
\end{description}

\subsection{Gaze Behavior}

Gaze behavior was kept constant across the two listening conditions, as direct eye contact was predicted to occur infrequently during the problem-solving sessions. We implemented the robot's gaze behavior based on the work by \citet{2001garau}. They found that adapting an avatar's gaze and head pose to the turn-taking pattern of a conversation outperformed a random gaze agent. The agent alternates between two actions, looking \textit{at} the user or \textit{away} from them (gaze aversion). The duration of each action is informed by previous work on dyadic interactions \cite{1976argyle,1967kendon} and varies depending on whether the agent is listening or speaking. For this work, we only used the listening timing indications. 


 
 



\section{Study Design}\label{sec:problem}


We designed a between-subject study with three conditions in order to evaluate the effects of different previously proposed robot listener behaviors in the outcome of a TAPS activity. Two conditions concerned robot listener behaviors: the first, a rule-based heuristic where non-verbal utterances and head nods are crafted according to previous literature (as described in Section \ref{sec:naivel}); the second, based on deep learning models which learn listener behaviors through data input from human-human dyadic interactions (as described in Section \ref{sec:datal}). The two listening behaviors were chosen to gauge if different complexity in the model used to generate the listener behavior affects the outcomes of TAPS activities. We compare these effects with the baseline condition, the presence of an inanimate object - a rubber duck - to whom words are directed. The three conditions are detailed in Section \ref{subsec:cond}.

\subsection{Hypotheses}

\label{subsec:hyp}

We were interested in evaluating how a robot can affect the think-aloud process in a "rubber duck debugging" session. Thus, we defined our outcomes based on how the user interacted with the task. We measure user \textit{engagement} in thinking-aloud and user \textit{perception} of the problem-solving session (or user experience). \textbf{Engagement} with the think-aloud process measures how the user talks when thinking aloud, e.g. number and duration of verbal utterances. 
\textbf{User experience} addresses how the participant perceives solving the task, including self-assessed cognitive load (or "mental effort") and enjoyment (or "user experience index"). These measures are detailed in Section \ref{subsec:metrics}.

\citet{2019wijnen} found that the presence of a robot leads to better explaining behavior while learning in children, and \citet{2018ramachandran} found that children engage more in thinking aloud utterances when a robot is present. Consequently, we hypothesized:
\begin{itemize}
	\item \textbf{H1a:} \textit{Engagement in thinking aloud is higher when thinking aloud to an actively listening robot than to an inanimate object.} 
\end{itemize}
Moreover, \citet{2010morency} report that a multimodal probabilistic model is better at predicting backchannels than hand-crafted rules. As such, we expected the data-driven condition to generate listener behavior that leads to a more organic interaction:
\begin{itemize}
	\item \textbf{H1b:} \textit{Listener behaviors learned from data lead to higher degrees of engagement in thinking-aloud, when compared to rule-based heuristics.}
\end{itemize}



Following \textbf{H1a, H1b}, and because the verbalization of one's thoughts during problem-solving has been reported to have affect how users solve a task \cite{1985stinessen,1994michelene}, we also hypothesize about user behavior during the task:
\begin{itemize}
    \item \textbf{H2a:} \textit{User behavior during task completion will be different in the conditions when a robot is present. Users will take longer to solve tasks, but also perform better.}
    \item \textbf{H2b:} \textit{Following \textbf{H1b}, users with the DL-based listener robot condition will take a longer time completing the tasks than users in the rule-based listener robot condition.} 

\end{itemize}

Finally, we wanted to evaluate the subjective experience of the user during the problem-solving sessions. Literature is sparse regarding if thinking aloud affects how users perceive the task, and if different listening behaviors skew that perception. As such, we took an exploratory approach to answer the following research questions: 
\begin{itemize}
	\item \textbf{RQ1a:} \textit{Is user experience affected by an active listener robot when compared to an inanimate object?}
	\item \textbf{RQ1b:} \textit{Is user experience affected by the type of listening behavior the robot exhibits?}
\end{itemize}

\subsection{Conditions}
\label{subsec:cond}

We sought to evaluate how the presence and behavior of a social robot impacted the outcome of a "rubber duck" problem-solving session. We designed a between-subject study with three conditions:
\begin{description}[align=left, leftmargin=0em, labelsep=0.2em, font=\textbf, itemsep=0em,parsep=0.3em]
\item[Rubber Duck (RDuck):] To reproduce a "rubber duck debugging" scenario, participants in the \textit{RDuck} condition were instructed to think aloud directing their words to a rubber duck.
\item[Naïve Listener Robot (NaiveL):] The social robot displayed listening behavior generated by hand-crafted heuristics. Information on the detailed implementation is given in Section \ref{sec:naivel}.
\item[Data-driven Listener Robot (DataL):] The social robot displayed listening behavior that was learned through machine learning methods using a human-human conversational corpus. The implementation details are provided in Section \ref{sec:datal}.

By including two different robot listening behaviors, we explored the transferability of two implementation methods of different complexity into a new social context, as well as the sensitivity of the user to differences between these behaviors.


\subsection{Problem-solving Tasks}
\label{subsec:task}

We designed two different tasks in order to evaluate distinct contexts for thinking-aloud problem-solving and how the presence of a robot can impact the perception of those tasks and the outcomes.

\subsubsection{\textbf{Logic Deduction Quiz (LogicQ)}} Tasks using deductive logic are common when developing thinking-aloud protocols \cite{2009leighton}. The questionnaire used here was adapted from \textit{IQ Test Labs}\footnote{https://www.intelligencetest.com/questions/logic/1.html}. For each question, 5 possible answers were shown as illustrated in Figure \ref{fig:logicd}. Participants were able to skip questions but could not see if their answers were correct. There was no time limit per question.

\subsubsection{\textbf{NASA Moon Survival Task (NASA)}}  In the NASA task \cite{1970hall}, the participant is placed in an hypothetical life-threatening scenario (spaceship crashed on the moon) and asked to rank a list of 15 available items according to their contribution towards survival. 

The LogicQ task allowed for an objective evaluation of performance by counting correct answers to the quiz, whereas the NASA task was framed as an open-ended question involving strategic thinking. We analyzed the effects of each task separately, as they elicit different cognitive processes, and tasks similar to these give rise to different protocols in thinking-aloud studies \cite{1994vansomeren}.

\begin{figure}
    \centering
    \includegraphics[width=0.25\textwidth]{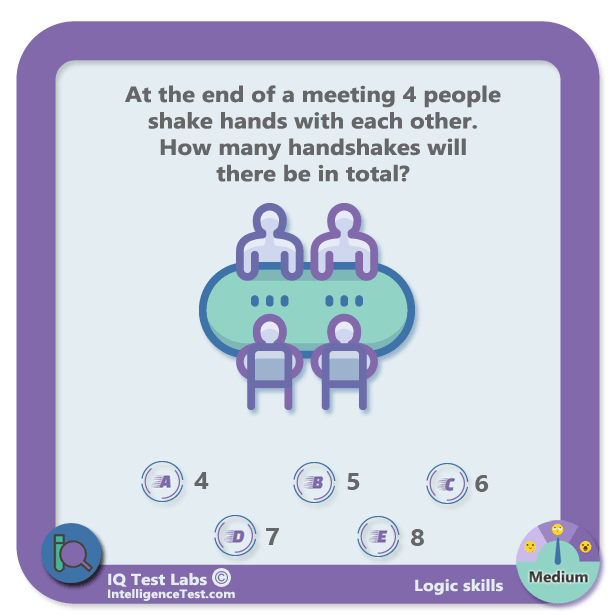}
        \caption{Layout example for one of the questions from the deductive logic quiz, from \textit{IQ Test Labs}.
    }
    \label{fig:logicd}
\end{figure}
\subsection{Measures} 
\label{subsec:metrics}
We established a set of metrics that explore the outcomes of the "rubber duck" problem-solving session. \textbf{Engagement} in thinking-aloud and \textbf{Task-related} metrics quantify the objective effects of the different conditions on how the user acted, whereas \textbf{User experience} attempts to quantify the subjective experience of the user while solving the tasks.

\subsubsection{Engagement metrics} To assess \textbf{\textit{H1}}, we measured the user's \textbf{Number} and \textbf{Duration of verbal utterances}, as well as \textbf{Speech-to-silence ratio} (time spent talking divided by time spent silent, during a task) while solving each task. This provided an objective measure of the user's engagement with the think-aloud process. This data was collected from Voice Activation Detection (VAD) through lapel microphones. 

\subsubsection{Task-related metrics} For evaluation of \textbf{\textit{H2}}, we defined metrics that relate to how the user behaved while solving the task. We measured performance in the \textit{LogicQ} task by monitoring the number of \textbf{Questions answered (QAns)}, and number of \textbf{Correct answers}. For the \textit{NASA} task, we considered \textbf{time spent} before submitting the final answer.

\subsubsection{User experience metrics} Finally, to investigate \textbf{\textit{RQ1}}, we measured self-reported \textbf{Cognitive load} and \textbf{User experience index (UEI)} of the problem solving session. This was assessed in a questionnaire taken \textit{after each task}. Cognitive load was measured using the NASA-TLX scale \cite{1988nasatlx}. The perception of the problem-solving session was evaluated with three dimensions from the short version of the User Engagement Scale (focused attention, perceived usability and reward) \cite{2018obrien}.

\subsubsection{Other measures}
We also collected data on participants' demographics (age, gender identity, country of origin), as well as the personality trait of \textbf{Extraversion} from the Big Five Inventory\cite{john1991big,2007rammstedt} in the pre-experiment questionnaire. In the post-experiment questionnaire, we assessed the \textbf{Impact of thinking aloud}, which asked users to rate how they thought thinking aloud helped them solve the tasks. This measure was obtained through five questions with a 5-point Likert scale. A Cronbach's Alpha of $\alpha=0.84$ indicated good reliability. The exact questions are provided as supplemental material. We further measured \textbf{Extraversion} and \textbf{Impact of thinking aloud} since \textit{personality} and \textit{natural inclination for speaking aloud} can impact user behavior during the tasks. Finally, we considered the \textbf{Order} of task presentation, since thinking aloud can suffer from a "cold start effect" \cite{1997gibson} (as described in Section \ref{subsec:exp}).

Finally, to explore if differences in user behavior could be due to the perception of the robot listening behaviors, the participants with conditions where the robot was present were asked about the robot's \textbf{Listening behavior} and \textbf{Closeness} \cite{2021murray}, \textbf{Social attributes} (from RoSAs \cite{rosas}), as well as previous experience with robots.

\subsection{Experimental procedure}
\label{subsec:exp}


Participants were randomly assigned to one of the three conditions. After giving informed consent, participants read the experiment instructions. In all conditions, they were instructed to speak aloud to "Alex", their "rubber duck", in a study that aims to understand cognitive processes behind problem-solving tasks. The instructions also clarified that Alex did not have any insight into the solutions of any of the two tasks. Afterwards, participants were asked to fill the pre-experiment questionnaire (demographic and personality information). Before moving to the experiment room, participants were given 2 minutes to respond to questions from the LogicQ task, in order to get familiar with the types of questions presented in the quiz.

Participants then moved to the experimental room. They sat at the desk, facing a tablet and "Alex". "Alex" was either the rubber duck or a Furhat robot, depending on the assigned condition. Figure \ref{fig:setup} illustrates the set-up. Audio features and voice activity were collected through a lapel microphone. Before starting the first task, participants were introduced to "Alex" and were asked to introduce themselves to it. This was prompted by the robot (or by the researcher in the \textit{RDuck} condition) and aimed to a) eliminate the "cold start effect' of talking aloud \cite{1997gibson} and, for the \textit{NaïveL} and \textit{DataL} conditions, b) eliminate the novelty effect \cite{2016kennedy} when first speaking to the robot. Participants were given $7$ minutes to complete each task (for the NASA task, they were free to submit their answer before that time). Between the two tasks, they were asked to fill out a questionnaire about the task they had just solved (collection of user experience measures). Task order was counterbalanced between participants, within each condition. After the second task, participants filled out the second task-based questionnaire and a final questionnaire (collection of perception of the robot and impact of thinking aloud measures).
After a final debriefing participants received a voucher valued at around 10 USD.

\paragraph{\textbf{System implementation}} As our social robot, we used a Furhat robot\footnote{https://furhatrobotics.com/} with the William voice from CereProc\footnote{https://cereproc.com/}, including its set of backchannels. For the nod movement, the "amplitude" - range of the up and down movement - was randomly sampled from a uniform distribution. During the nod, the robot paused at the lowest point for $0.5$ s.  Furhat was controlled through a computer using a NVIDIA GeForce RTX 2080 SUPER and an Intel® Core™ i9.

\paragraph{\textbf{Participants}}

Participants were recruited through posters, flyers, social media platforms and word of mouth. A total of 101 participants were recruited with ages ranging from 14-76 years (M = 26.4, SD = 7.6). 53 participants identified as male and 48 as female, with a total of 29 different nationalities. 
\end{description}

\section{Results} \label{sec:results}

Data from eleven participants had to be excluded due to substantial software or hardware problems. The remaining 90 participants were distributed across conditions with the following demographics: $RDuck - N = 30$ $(14F,16M)$, ages $ 27.8 \pm 10.2$; $NaiveL - N = 29$ $(9F,20M)$, ages $26.3 \pm 7.4$; $DataL - N = 31$ $(18F,13M)$, ages $25.3 \pm 5.3$. Out of the 60 participants which interacted with the robot, 20 indicated they had interacted with a robot before. In this section, we analyze the listening behaviors developed and report our findings for each measure presented in Section \ref{sec:problem}.

\subsection{User Engagement in Thinking Aloud}

We evaluated how much participants engaged in thinking aloud utterances in two separate tasks with an ANCOVA to examine the effects of \textit{condition} on the \textbf{Number} and \textbf{Duration of utterances}, as well as \textbf{Speech-to-silence ratio}, after controlling for \textbf{Extraversion} and \textbf{Impact of Thinking Aloud}. We also control for the \textbf{Order} of presentation of tasks, due to the "cold start effect" mentioned in Section \ref{sec:problem} \cite{1997gibson}. Table \ref{tab:vad} summarizes our findings.

\subsubsection{NASA task}

The effect of the \textit{condition} on the \textbf{Number of utterances} was not significant ($F(2,83)=0.33,p=0.71$). The covariate \textbf{Order} was a significant predictor for the \textbf{Duration of utterances} and \textbf{Speech-to-silence ratio} ( $F(1,83)=4.16,p=0.04$ and $F(1,83)=5.84, p=0.02$, resp.). Doing the NASA task first decreased mean \textbf{duration of utterances} ($3.30\pm1.31 s$ vs. $3.72\pm1.85 s$ if doing the LogicQ task first) and \textbf{speech-to-silence ratio} ($1.46\pm1.01$ vs. $1.88\pm1.74$). 

\subsubsection{LogicQ task}
The ANCOVA test showed no significant effects of the \textit{condition} on the engagement of the user with the task (\textbf{Number of utterances}, $F(2,83)=0.21,p=0.81$, \textbf{Duration of utterances}, $F(2,83)=1.15,p=0.32$, \textbf{Speech-to-silence ratio}, $F(2,83)=1.18,p=0.31$). 

\begin{table}
\centering
\tabcolsep=0.11cm
\caption{User Engagement metrics, for both tasks ($M\pm SD$).}
\label{tab:vad}
\begin{tabular}{c|c|c|c|c}
\hline
\textbf{Task}                   & \textbf{Condition} & \textit{\textbf{Number}} & \textit{\textbf{Duration}} & \textit{\textbf{Speech-to-silence}} \\ \hline
\multirow{3}{*}{\textbf{NASA}}  & \textit{RDuck}     & $0.17\pm0.04$            & $3.50\pm1.34$              & $1.73\pm1.12$                       \\
                                & \textit{NaiveL}    & $0.16\pm0.04$            & $3.70\pm1.46$              & $1.73\pm1.19$                       \\
                                & \textit{DataL}     & $0.16\pm0.04$            & $4.37\pm2.66$              & $2.55\pm2.44$                       \\ \hline
\multirow{3}{*}{\textbf{LogicQ}} & \textit{RDuck}     & $0.16\pm0.05$            & $2.90\pm0.87$              & $1.21\pm0.78$                       \\
                                & \textit{NaiveL}    & $0.16\pm0.04$            & $3.28\pm1.32$              & $1.27\pm0.85$                       \\
                                & \textit{DataL}     & $0.17\pm0.03$            & $3.31\pm1.10$              & $1.52\pm1.11$                       \\
\end{tabular}
\end{table}

\subsection{Task-related metrics}

Table \ref{tab:performance} shows task-related measures across participants within each condition. For the NASA task, we consider \textbf{Time taken} until submission of final object ranking (in seconds, with a maximum of $420 s$). For the LogicQ task, we looked at the \textbf{Number of Questions Answered (QAns)} and \textbf{Number of Correct Answers} as a way to measure how the different conditions affected how the user completes the LogicQ task. We controlled for the \textbf{Impact of Thinking Aloud} in both tasks. 

\subsubsection{NASA task}

An ANCOVA revealed no significant effect of \textit{condition} on \textbf{Time Taken} solving the task ($F(2,84) = 1.67, p=0.19$).

\subsubsection{LogicQ task}
A one-way ANOVA revealed no significant effect of \textit{condition} on \textbf{Correct answers} ($F(2,84) = 0.21, p=0.81$), or on \textbf{Number of questions answered} ($F(2,84) = 0.14, p=0.86$).

\begin{table}
\centering
\tabcolsep=0.11cm
\caption{Task Performance metrics, for both tasks ($M\pm SD$).}
\label{tab:performance}
\begin{tabular}{c|c|c|c|c}
\hline
\textbf{Task}                   & \textbf{Condition} & \textit{\textbf{QAns}} & \textit{\textbf{Correct}} & \textit{\textbf{Time Taken (s)}} \\ \hline
\multirow{3}{*}{\textbf{NASA}}  & \textit{RDuck}     & -                      & -                             & $360\pm75$                       \\
                                & \textit{NaiveL}    & -                      & -                             & $349\pm80$                       \\
                                & \textit{DataL}     & -                      & -                             & $320\pm99$                       \\ \hline
\multirow{3}{*}{\textbf{LogicQ}} & \textit{RDuck}     & $8.41\pm2.93$         & $0.011\pm0.005$                & -                                \\
                                & \textit{NaiveL}    & $8.93\pm3.50$         & $0.010\pm0.36$                & -                                \\
                                & \textit{DataL}     & $8.58\pm3.50$         & $0.009\pm0.005$                & -                               
\end{tabular}
\end{table}

\subsection{User Experience metrics}
\begin{table}
\centering
\tabcolsep=0.11cm
\caption{User Experience metrics, for both tasks ($M\pm SD$).}
\label{tab:ues}
\begin{tabular}{c|c|c|c}
\hline
\textbf{Task}                   & \textbf{Condition} & \textit{\textbf{Cog. Load}} & \multicolumn{1}{c}{\textit{\textbf{UEI}}} \\ \hline
\multirow{3}{*}{\textbf{NASA}}  & \textit{RDuck}     & $3.47\pm0.87$               & $3.42\pm0.31$                               \\
                                & \textit{NaiveL}    & $3.53\pm0.73$               & $3.39\pm0.31$                              \\
                                & \textit{DataL}     & $3.11\pm0.67$               & $3.40\pm0.33$                              \\ \hline
\multirow{3}{*}{\textbf{LogicQ}} & \textit{RDuck}     & $4.18\pm0.88$               & $3.40\pm0.29$                               \\
                                & \textit{NaiveL}    & $4.08\pm0.81$               & $3.26\pm0.30$                              \\
                                & \textit{DataL}     & $4.11\pm0.84$               & $3.27\pm0.34$                             
\end{tabular}
\end{table}
Participants' self-reported experience solving the tasks was evaluated through measures of \textbf{Cognitive load} and \textbf{User Experience Index} ($UEI$). We tested the effect of condition on these metrics with an ANCOVA, controlling for the \textbf{Order} of tasks and \textbf{Impact of Thinking Aloud}. See Table \ref{tab:ues} for more detail.

\subsubsection{NASA task}
The effect of the \textit{condition} on self-reported \textbf{Cognitive load} is not significant, $F(2,84)=2.43, p=0.09$.
The covariate \textbf{Order} of task presentation significantly effected (decreased) \textbf{UEI}, $F(1,84)=8.11,p=0.005$ (NASA first, \textbf{UEI} $=3.32\pm0.31$; LogicQ first, \textbf{UEI} $=3.39\pm0.32$).


\subsubsection{LogicQ task}

A one-way ANCOVA showed no significant effects of the \textit{condition} on User Experience metrics (\textbf{Cognitive Load}, $F(2,86)=0.05,p=0.95$, \textbf{UEI}, $F(2,86)=2.05,p=0.14$).


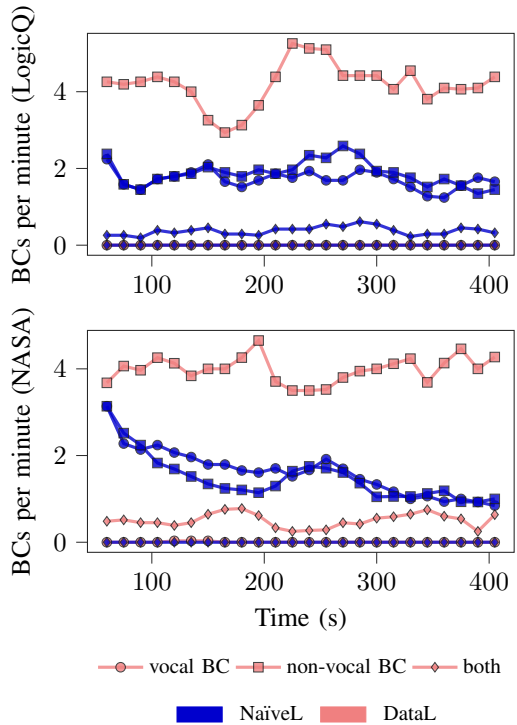
\begin{figure}[t]
    \centering


\begin{tikzpicture}
\definecolor{black25}{RGB}{25,25,25}
\definecolor{color0}{rgb}{0.392156862745098,0.584313725490196,0.929411764705882}
\definecolor{color1}{rgb}{0.803921568627451,0.36078431372549,0.36078431372549}
\definecolor{color2}{rgb}{0.254901960784314,0.411764705882353,0.882352941176471}
\definecolor{color3}{rgb}{0.941176470588235,0.501960784313725,0.501960784313725}
\definecolor{color4}{rgb}{0.698039215686274,0.133333333333333,0.133333333333333}

\definecolor{ncolor0}{RGB}{0,0,205}
\definecolor{ncolor1}{rgb}{0.941176470588235,0.501960784313725,0.501960784313725}

\begin{groupplot}[group style={group size=1 by 2},width=0.4\textwidth, height=0.25\textwidth, legend style={draw=none,nodes={scale=0.8, transform shape}},]
\nextgroupplot[
axis line style={white!15!black},
tick align=outside,
tick pos=left,
x grid style={white!80!black},
xmin=42.75, xmax=422.25,
xtick style={color=white!15!black},
y grid style={white!80!black},
ylabel={BCs per minute (LogicQ)},
ymin=-0.262903225806452, ymax=5.52096774193548,
ytick style={color=white!15!black}
]
\addplot [very thick, ncolor0, opacity=0.8, mark=*, mark size=1.8, mark options={solid,draw=black25,line width=0.4pt}]
table {
60 2.24137931034483
75 1.58620689655172
90 1.44827586206897
105 1.72413793103448
120 1.79310344827586
135 1.89655172413793
150 2.10344827586207
165 1.6551724137931
180 1.51724137931034
195 1.68965517241379
210 1.86206896551724
225 1.75862068965517
240 1.93103448275862
255 1.68965517241379
270 1.68965517241379
285 1.96551724137931
300 1.89655172413793
315 1.72413793103448
330 1.51724137931034
345 1.27586206896552
360 1.24137931034483
375 1.55172413793103
390 1.75862068965517
405 1.6551724137931
};
\addplot [very thick, color1, opacity=0.8, mark=*, mark size=1.8, mark options={solid,draw=black25,line width=0.4pt}]
table {
60 0
75 0
90 0
105 0
120 0
135 0
150 0
165 0
180 0
195 0
210 0
225 0
240 0
255 0
270 0
285 0
300 0
315 0
330 0
345 0
360 0
375 0
390 0
405 0
};
\addplot [very thick, ncolor0, opacity=0.8, mark=square*, mark size=1.8, mark options={solid,draw=black25,line width=0.4pt}]
table {
60 2.37931034482759
75 1.58620689655172
90 1.44827586206897
105 1.72413793103448
120 1.79310344827586
135 1.86206896551724
150 2.03448275862069
165 1.89655172413793
180 1.79310344827586
195 1.96551724137931
210 1.86206896551724
225 1.96551724137931
240 2.3448275862069
255 2.27586206896552
270 2.58620689655172
285 2.37931034482759
300 1.93103448275862
315 1.89655172413793
330 1.75862068965517
345 1.51724137931034
360 1.72413793103448
375 1.55172413793103
390 1.3448275862069
405 1.44827586206897
};
\addplot [very thick, ncolor1, opacity=0.8, mark=square*, mark size=1.8, mark options={solid,draw=black25,line width=0.4pt}]
table {
60 4.25806451612903
75 4.19354838709677
90 4.25806451612903
105 4.38709677419355
120 4.25806451612903
135 4
150 3.25806451612903
165 2.93548387096774
180 3.12903225806452
195 3.64516129032258
210 4.38709677419355
225 5.25806451612903
240 5.12903225806452
255 5.09677419354839
270 4.41935483870968
285 4.41935483870968
300 4.41935483870968
315 4.06451612903226
330 4.54838709677419
345 3.80645161290323
360 4.09677419354839
375 4.06451612903226
390 4.09677419354839
405 4.38709677419355
};
\addplot [very thick, blue!80.3921568627451!black, opacity=0.8, mark=diamond*, mark size=1.8, mark options={solid,draw=black25,line width=0.4pt}]
table {%
60 0
75 0
90 0
105 0
120 0
135 0
150 0
165 0
180 0
195 0
210 0
225 0
240 0
255 0
270 0
285 0
300 0
315 0
330 0
345 0
360 0
375 0
390 0
405 0
};
\addplot [very thick, ncolor0, opacity=0.8, mark=diamond*, mark size=1.8, mark options={solid,draw=black25,line width=0.4pt}]
table {
60 0.258064516129032
75 0.258064516129032
90 0.193548387096774
105 0.387096774193548
120 0.32258064516129
135 0.387096774193548
150 0.451612903225806
165 0.290322580645161
180 0.290322580645161
195 0.258064516129032
210 0.419354838709677
225 0.419354838709677
240 0.419354838709677
255 0.548387096774194
270 0.483870967741935
285 0.612903225806452
300 0.548387096774194
315 0.387096774193548
330 0.225806451612903
345 0.290322580645161
360 0.290322580645161
375 0.451612903225806
390 0.419354838709677
405 0.32258064516129
};
\vspace{-0.9cm}
\nextgroupplot[
axis line style={white!15!black},
legend cell align={left},
legend style={
  fill opacity=0.8,
  text opacity=1,
  at={(0.91,0.5)},
  anchor=east,
  draw=none,
  legend style={at={($(0,0)+(1cm,1cm)$)},legend columns=4,fill=none,
  anchor=center,align=center},
            legend to name=fred
},
tick align=outside,
tick pos=left,
x grid style={white!80!black},
xlabel={Time (s)},
xmin=42.75, xmax=422.25,
xtick style={color=white!15!black},
y grid style={white!80!black},
ylabel={BCs per minute (NASA)},
ymin=-0.232692307692308, ymax=4.88653846153846,
ytick style={color=white!15!black}
]
\addplot [very thick, ncolor0, opacity=0.8, mark=*, mark size=1.8, mark options={solid,draw=black25,line width=0.4pt}, forget plot]
table {
60 3.13793103448276
75 2.27586206896552
90 2.13793103448276
105 2.24137931034483
120 2.06896551724138
135 1.96551724137931
150 1.79310344827586
165 1.79310344827586
180 1.6551724137931
195 1.60714285714286
210 1.7037037037037
225 1.52
240 1.66666666666667
255 1.91666666666667
270 1.69565217391304
285 1.45454545454545
300 1.33333333333333
315 1.16666666666667
330 1
345 1.0625
360 0.9375
375 1
390 0.928571428571429
405 0.846153846153846
};
\addplot [very thick, ncolor1, opacity=0.8, mark=*, mark size=1.8, mark options={solid,draw=black25,line width=0.4pt}]
table {
60 0
75 0
90 0
105 0
120 0.032258064516129
135 0.032258064516129
150 0.032258064516129
165 0
180 0
195 0
210 0
225 0
240 0
255 0
270 0
285 0
300 0
315 0
330 0
345 0
360 0
375 0
390 0
405 0
};
\addlegendentry{vocal BC}
\addplot [very thick, ncolor0, opacity=0.8, mark=square*,mark size=1.8, mark options={solid,draw=black25,line width=0.4pt}, forget plot]
table {
60 3.13793103448276
75 2.51724137931034
90 2.24137931034483
105 1.82758620689655
120 1.68965517241379
135 1.51724137931034
150 1.3448275862069
165 1.24137931034483
180 1.20689655172414
195 1.14285714285714
210 1.2962962962963
225 1.64
240 1.75
255 1.70833333333333
270 1.60869565217391
285 1.36363636363636
300 1.04761904761905
315 1.05555555555556
330 1.05555555555556
345 1.125
360 1.1875
375 0.933333333333333
390 0.928571428571429
405 1
};
\addplot [very thick, ncolor1, opacity=0.8, mark=square*, mark size=1.8, mark options={solid,draw=black25,line width=0.4pt}]
table {
60 3.67741935483871
75 4.06451612903226
90 3.96774193548387
105 4.25806451612903
120 4.12903225806452
135 3.83870967741935
150 4
165 4
180 4.25925925925926
195 4.65384615384615
210 3.70833333333333
225 3.5
240 3.5
255 3.52380952380952
270 3.8
285 3.94736842105263
300 4
315 4.11764705882353
330 4.23529411764706
345 3.6875
360 4.13333333333333
375 4.46153846153846
390 4
405 4.27272727272727
};
\addlegendentry{non-vocal BC}
\addplot [very thick, ncolor0, opacity=0.8, mark=diamond*, mark size=1.8, mark options={solid,draw=black25,line width=0.4pt}, forget plot]
table {
60 0
75 0
90 0
105 0
120 0
135 0
150 0
165 0
180 0
195 0
210 0
225 0
240 0
255 0
270 0
285 0
300 0
315 0
330 0
345 0
360 0
375 0
390 0
405 0
};
\addplot [very thick, ncolor1, opacity=0.8, mark=diamond*, mark size=1.8, mark options={solid,draw=black25,line width=0.4pt}]
table {
60 0.483870967741935
75 0.516129032258065
90 0.451612903225806
105 0.451612903225806
120 0.387096774193548
135 0.451612903225806
150 0.645161290322581
165 0.758620689655172
180 0.777777777777778
195 0.615384615384615
210 0.333333333333333
225 0.25
240 0.272727272727273
255 0.285714285714286
270 0.45
285 0.421052631578947
300 0.555555555555556
315 0.588235294117647
330 0.647058823529412
345 0.75
360 0.6
375 0.538461538461538
390 0.25
405 0.636363636363636
};
\addlegendentry{both}
\coordinate (c1) at (rel axis cs:0.5,0.5);
\coordinate (c2) at (rel axis cs:0.85,0.85);
\end{groupplot}
\node[below] at (c1 |- current bounding box.south)
      {\pgfplotslegendfromname{fred}};

\begin{customlegend}[legend columns=-1, 
legend cell align=center, 
legend entries={ 
NaïveL,
DataL
},
legend style={draw=none,column sep=1ex,at={(c2 |- current bounding box.south)}, font=\footnotesize}] 

    \addlegendimage{area legend,ncolor0,fill = ncolor0} 
    \addlegendimage{area legend,ncolor1,fill=ncolor1} 
\end{customlegend}

\end{tikzpicture}
    \caption{Backchanneling behavior over time from the DL-based model (\textit{DataL}, in red) or heuristic model (\textit{NaïveL}, in blue). The different types of BCs emmited are discriminated. Data shown is averaged over all participants in that condition. Top - LogicQ task; bottom - NASA task.}
    \label{fig:robotbehavior}
\end{figure}

\subsection{Robot Behavior}

In addition to the main measures, we also looked into how the robot acted and how it was perceived in the \textit{NaïveL} and \textit{DataL} conditions. This helps to illustrate the two listener behavior models implemented, but can also correlate with user behavior during the TAPS sessions under different conditions.

Backchannel frequency per minute (see Fig. \ref{fig:robotbehavior}) is given by the count of backchannels executed by the robot every 60 seconds. It is calculated as a sliding window with hop length of $15 s$.
The \textit{NaïveL} robot displayed similar values for vocal and non-vocal backchanneling, but the frequency decreased as task time increases. The \textit{DataL} robot, on the other hand, favors non-vocal backchannels (nodding), which is displayed in much higher frequency, as well as a combination of both nodding and vocal utterances.

\subsubsection{User Reporting of Robot Behavior}

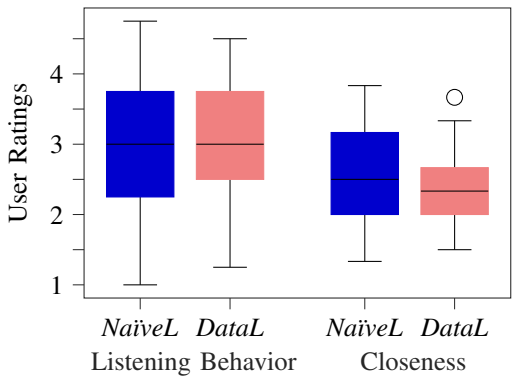
\begin{figure}
     \centering

\begin{tikzpicture}

\definecolor{color0}{rgb}{0.254901960784314,0.411764705882353,0.882352941176471}
\definecolor{color1}{rgb}{0.941176470588235,0.501960784313725,0.501960784313725}
\definecolor{color0}{RGB}{0,0,205}
\definecolor{color1}{rgb}{0.941176470588235,0.501960784313725,0.501960784313725}

\begin{axis}[
axis line style={white!15!black},
tick align=outside,
tick pos=left,
x grid style={white!80!black},
xmin=-0.4, xmax=3.4,
xtick style={color=white!15!black},
xtick={0.1,0.9,2.1,2.9},
xticklabels={\textit{NaïveL},\textit{DataL},\textit{NaïveL},\textit{DataL}},
y grid style={white!80!black},
ylabel={User Ratings},
ymin=0.8125, ymax=4.9375,
ytick style={color=white!15!black},
ytick={1,1.5,2,2.5,3,3.5,4,4.5,5},
yticklabels={1,,2,,3,,4,,5},
width=0.4\textwidth, height=0.3\textwidth, clip=false
]
\addplot [black]
table {%
0.1 2.25
0.1 1
};
\addplot [black]
table {%
0.1 3.75
0.1 4.75
};
\addplot [black]
table {%
-0.05 1
0.25 1
};
\addplot [black]
table {%
-0.05 4.75
0.25 4.75
};
\addplot [black]
table {%
0.9 2.5
0.9 1.25
};
\addplot [black]
table {%
0.9 3.75
0.9 4.5
};
\addplot [black]
table {%
0.75 1.25
1.05 1.25
};
\addplot [black]
table {%
0.75 4.5
1.05 4.5
};
\addplot [black]
table {%
2.1 2
2.1 1.33333333333333
};
\addplot [black]
table {%
2.1 3.16666666666667
2.1 3.83333333333333
};
\addplot [black]
table {%
1.95 1.33333333333333
2.25 1.33333333333333
};
\addplot [black]
table {%
1.95 3.83333333333333
2.25 3.83333333333333
};
\addplot [black]
table {%
2.9 2
2.9 1.5
};
\addplot [black]
table {%
2.9 2.66666666666667
2.9 3.33333333333333
};
\addplot [black]
table {%
2.75 1.5
3.05 1.5
};
\addplot [black]
table {%
2.75 3.33333333333333
3.05 3.33333333333333
};
\addplot [black, mark=o, mark size=3, mark options={solid,fill opacity=0}, only marks]
table {%
2.9 3.66666666666667
};
\path [draw=color0, fill=color0]
(axis cs:-0.2,2.25)
--(axis cs:0.4,2.25)
--(axis cs:0.4,3.75)
--(axis cs:-0.2,3.75)
--(axis cs:-0.2,2.25)
--cycle;
\path [draw=color1, fill=color1]
(axis cs:0.6,2.5)
--(axis cs:1.2,2.5)
--(axis cs:1.2,3.75)
--(axis cs:0.6,3.75)
--(axis cs:0.6,2.5)
--cycle;
\path [draw=color0, fill=color0]
(axis cs:1.8,2)
--(axis cs:2.4,2)
--(axis cs:2.4,3.16666666666667)
--(axis cs:1.8,3.16666666666667)
--(axis cs:1.8,2)
--cycle;
\path [draw=color1, fill=color1]
(axis cs:2.6,2)
--(axis cs:3.2,2)
--(axis cs:3.2,2.66666666666667)
--(axis cs:2.6,2.66666666666667)
--(axis cs:2.6,2)
--cycle;
\addplot [black]
table {%
-0.2 3
0.4 3
};
\addplot [black]
table {%
0.6 3
1.2 3
};
\addplot [black]
table {%
1.8 2.5
2.4 2.5
};
\addplot [black]
table {%
2.6 2.33333333333333
3.2 2.33333333333333
};
\draw (axis cs:-0.42, -0.12) node[
  anchor=west,
  text=white!15!black,
  rotate=0.0
]{Listening Behavior};
\draw (axis cs:1.98,-0.08) node[
  anchor=west,
  text=white!15!black,
  rotate=0.0
]{Closeness};
\end{axis}

\end{tikzpicture}
    \caption{Participants' rating of the robot's listening behavior and closeness to the robot (from \cite{2021murray}).}
   
    \label{fig:robotbehavior_listening}
\end{figure}

Fig. \ref{fig:robotbehavior_listening} shows participants' evaluation of the robot's listening behavior. A one-way ANOVA showed no significant differences between conditions for both the \textbf{Closeness} and \textbf{Listening behavior} dimensions ($F(1,58)=2.07,p=0.16$ and $F(1,58)=0.56, p=0.46$, respectively). The robot's social attributes \cite{rosas} were also not significantly different between the \textit{NaiveL} and \textit{DataL} conditions (one-way ANOVA for the \textbf{Competence} dimension, $F(1,58)=0.46,p=0.50$, and Wilcoxon rank sum test for the \textbf{Warmth}, $W=543.5,p=0.10$ and \textbf{Discomfort}, $W=410,p=0.55$, dimensions, after a Shapiro–Wilk test of normality revealed these variables did not have normal distributions, $p<0.05$).

\section{Discussion} \label{sec:discussion}

In order to capture the impact of a "robot duck," we analyzed various aspects of a thinking-aloud problem-solving session. Overall, we cannot say that the presence of the robot impacted the "rubber duck" TAPS sessions, nor that the two listening behavior models caused significant differences in task performance or user experience. Our results reject \textbf{H1a,b} - the presence of the robot, in either condition (\textit{NaïveL} or \textit{DataL}), did not appear to affect user engagement in thinking aloud. This result is in contrast to the findings in \citet{2018ramachandran}, where children thinking aloud to a robot spoke more than those in a no-robot condition. We also reject \textbf{H2a,b} - task performance was not affected, and the time spent to complete the tasks did not significantly change regardless of the condition. 
Finally, \textbf{RQ1} was aimed at the impact of the presence of the robot on subjective user experience. \citet{2013jung} found that users in human-robot teams with robots that backchannel show lower cognitive load when solving tasks. However, we found no significant effects of condition on how users self-reported the sessions. Below, we advance potential explanations for the findings in this study.

The data-driven model appeared to greatly favor non-verbal feedback, with rare instances of vocal-only backchannels. 
However, notably, the users rank both behaviors similarly in the dimensions of \textbf{Listening behavior}, \textbf{Closeness} and social attributes (\textbf{Competence, Warmth} and \textbf{Discomfort}). These findings differ from those reported in \citet{2021murray}, which found that a data-driven listener behavior was perceived as a better listener than a rule-based model. This might indicate that the user becomes fully involved with the task, paying little attention to the robot. This phenomenon has been observed in a setting where children playing a game stop listening to the robot's suggestions after some time \cite{2022ali}.

High user involvement in task completion could also explain why we do not find significant differences between the \textit{RDuck} condition and the two robot conditions. Relevant prior work on human dishonesty in the presence of a robot \cite{2022petisca} found that there were no differences in how users cheated between being alone or in the presence of an unaware robot. We hypothesize that there was a mismatch between the expectations of the users and robot behavior, leading users to believe that the robot did not truly understand what they were saying. Furhat's anthropomorphic appearance elicits the development of expectations about the system's performance \cite{2000nass}, causing cognitive anthropomorphization \cite{2022sacino}. Previous studies also showed different reactions to robot failure depending on the robot's appearance \cite{2020kontogiorgos} and functionality framing \cite{2020washburn}.
Inadequate backchannel timing thus may have had a detrimental effect on the expectations of the user, leading them to believe Furhat was unaware of the meaning of their utterances - and diluting the differences between the three conditions tested. This is consistent with comments left by the users in the post-experiment questionnaire, e.g., \textit{" (...) the robot is responsive however, can't replace a human interaction as I didn't get the emotional feedback.”}(P72).

The order of presentation of tasks is a significant predictor of measures such as \textbf{Speech-to-silence ratio} and self-reported \textbf{User Experience}. As mentioned in Section \ref{sec:problem}, an initial inhibition is to be expected when thinking aloud \cite{1997gibson}. Further, some participants reported that the two tasks differed in how organic it felt to say their thoughts aloud, with many feeling that it was harder to do it during \textit{LogicQ} quiz. Deductive logic skills also vary among humans, and this may explain why the \textbf{Impact of thinking aloud} is a predictor of involvement measures for the \textit{LogicQ} task and not the \textit{NASA} task. 

On a final note, we observe that much work in the development of robot listener behavior requires the collection of specific conversational data in chosen human contexts that are then replicated in human-robot interactions \cite{2010morency, 2019huang,2021murray}. Given the wide range of social settings and contextual characteristics within which HRI can be applied, one might wonder: is it sustainable to continue leveraging social HRI research by collecting task-specific datasets? In this work, we developed two fully-autonomous listening behaviors models that were informed by human-human conversational data collected in a contrasting social setting to that in which they were deployed. In spite of this, both behaviors lead to successful interactions of the user with the robot, with no significant detriment in any of the aspects of the sessions when compared to an inactive listener. This opens the door to continue exploring transferability of datasets to new social settings.

\subsection{Limitations and Future Work}
\label{sec:limit}

This study carries some limitations. Functionality misframing, or mismatch of user expectations towards the robot, could be reduced by providing participants with more interaction time with the robot before task completion, in order to adjust expectations. Since participants spent most of the time looking at the tablet, some reactions from the robot may have been missed and impacted the perception of robot behavior. Further, some participants described the robot as "intimidating", which could have impacted their comfort in thinking aloud.

Future work may help elucidate some questions raised by the results reported here. Age and cultural background may impact user behavior, and these aspects were not taken into account. 
There is also potential to explore if different-looking social robots affect the sessions, or the impact of a human listener. It would be interesting to further investigate if the outcomes of the TAPS session change with a different set of tasks (e.g. programming or debugging tasks).

Regarding listener behavior implementation, we did not explore multimodal data input (such as facial expressions, gaze tracking or head pose). Future work may explore these aspects, in line with what was developed in \cite{2019huang,2021murray}. The robot could also display more complex listening behavior, such as asking general questions to the user (e.g.\citet{1966weizenbaum}'s ELIZA). Finally, and while we explored transferability of datasets when implementing robot listener behavior, we note that behavior expectations may differ in a human-human versus a human-robot interaction \cite{moshkina2014social,gambino2020building}. This is yet another aspect that calls for further exploration.

\section{Conclusion} \label{sec:conclusion}


This study explored human-robot interaction in a new setting and broadened knowledge on the role and usefulness (or lack thereof) of social robots as adjacent tools for human task completion. Further, the development and/or implementation of two fully-autonomous robot listener behaviors provided an interesting contribution on the transferability of heuristics and conversational datasets into new interactive contexts. Our findings indicate that the presence of a social robot displaying elementary listener behavior is not sufficient to elicit substantial differences in human behavior and perception of a "rubber duck" think-aloud problem-solving session, when compared to an inanimate object. Additional studies are required to inform the design of socially-assistive robots in problem-solving - as the future brings us closer to social robots, an optimized "robot duck" could play an important role in the problem-solving tasks which populate our days.

\bibliographystyle{plainnat}
\balance
\bibliography{newbib.bib}

\begin{thebibliography}{58}
\providecommand{\natexlab}[1]{#1}
\providecommand{\url}[1]{\texttt{#1}}
\expandafter\ifx\csname urlstyle\endcsname\relax
  \providecommand{\doi}[1]{doi: #1}\else
  \providecommand{\doi}{doi: \begingroup \urlstyle{rm}\Url}\fi

\bibitem[Ali et~al.(2022)Ali, Devasia, and Breazeal]{2022ali}
Safinah Ali, Nisha~Elizabeth Devasia, and Cynthia Breazeal.
\newblock Escape!bot: Social robots as creative problem-solving partners.
\newblock In \emph{Creativity and Cognition}, C\&C '22, page 275–283, New
  York, NY, USA, 2022. Association for Computing Machinery.
\newblock ISBN 9781450393270.
\newblock \doi{10.1145/3527927.3532793}.
\newblock URL \url{https://doi.org/10.1145/3527927.3532793}.

\bibitem[Argyle and Cook(1976)]{1976argyle}
Michael Argyle and Mark Cook.
\newblock \emph{Gaze and mutual gaze}.
\newblock Cambridge University Press, 1976.
\newblock ISBN 0-521-20865-3.

\bibitem[Aubrey et~al.(2013)Aubrey, Marshall, Rosin, Vandeventer, Cunningham,
  and Wallraven]{ccdb}
Andrew~J. Aubrey, David Marshall, Paul~L. Rosin, Jason Vandeventer, Douglas~W.
  Cunningham, and Christian Wallraven.
\newblock Cardiff conversation database (ccdb): A database of natural dyadic
  conversations.
\newblock In \emph{2013 IEEE Conference on Computer Vision and Pattern
  Recognition Workshops}, pages 277--282, 2013.
\newblock \doi{10.1109/CVPRW.2013.48}.

\bibitem[Busso et~al.(2008)Busso, Bulut, Lee, Kazemzadeh, Mower, Kim, Chang,
  Lee, and Narayanan]{iemocap}
Carlos Busso, Murtaza Bulut, Chi-Chun Lee, Abe Kazemzadeh, Emily Mower, Samuel
  Kim, Jeannette~N. Chang, Sungbok Lee, and Shrikanth~S. Narayanan.
\newblock {IEMOCAP}: interactive emotional dyadic motion capture database.
\newblock \emph{Language Resources and Evaluation}, 42\penalty0 (4):\penalty0
  335--359, December 2008.

\bibitem[Butchart et~al.(2021)Butchart, Harrison, Ritchie, Mart{\'\i},
  McCarthy, Knight, and Scheinberg]{butchart2021child}
Joanna Butchart, Reema Harrison, Jan Ritchie, Felip Mart{\'\i}, Chris McCarthy,
  Sarah Knight, and Adam Scheinberg.
\newblock Child and parent perceptions of acceptability and therapeutic value
  of a socially assistive robot used during pediatric rehabilitation.
\newblock \emph{Disability and rehabilitation}, 43\penalty0 (2):\penalty0
  163--170, 2021.

\bibitem[Carpinella et~al.(2017)Carpinella, Wyman, Perez, and
  Stroessner]{rosas}
Colleen~M. Carpinella, Alisa~B. Wyman, Michael~A. Perez, and Steven~J.
  Stroessner.
\newblock {The Robotic Social Attributes Scale (RoSAS)}.
\newblock \emph{Proceedings of the 2017 ACM/IEEE International Conference on
  Human-Robot Interaction - HRI '17}, \penalty0 (October):\penalty0 254--262,
  2017.
\newblock ISSN 21672148.
\newblock \doi{10.1145/2909824.3020208}.
\newblock URL \url{http://dl.acm.org/citation.cfm?doid=2909824.3020208}.

\bibitem[Casas et~al.(2021)Casas, Senft, Gutierrez, Rincon-Rocancio, Munera,
  Belpaeme, and Cifuentes]{casas2021social}
Jonathan Casas, Emmanuel Senft, Luisa~F Gutierrez, Monica Rincon-Rocancio,
  Marcela Munera, Tony Belpaeme, and Carlos~A Cifuentes.
\newblock Social assistive robots: assessing the impact of a training assistant
  robot in cardiac rehabilitation.
\newblock \emph{International Journal of Social Robotics}, 13\penalty0
  (6):\penalty0 1189--1203, 2021.

\bibitem[Chi et~al.(1994)Chi, {De Leeuw}, Chiu, and Lavancher]{1994michelene}
Michelene~T.H. Chi, Nicholas {De Leeuw}, Mei-Hung Chiu, and Christian
  Lavancher.
\newblock Eliciting self-explanations improves understanding.
\newblock \emph{Cognitive Science}, 18\penalty0 (3):\penalty0 439--477, 1994.
\newblock ISSN 0364-0213.
\newblock \doi{https://doi.org/10.1016/0364-0213(94)90016-7}.
\newblock URL
  \url{https://www.sciencedirect.com/science/article/pii/0364021394900167}.

\bibitem[{de Kok} and Heylen(2012)]{2012dekok}
I.A. {de Kok} and {Dirk K.J.} Heylen.
\newblock A survey on evaluation metrics for backchannel prediction models.
\newblock In \emph{Proceedings of the Interdisciplinary Workshop on Feedback
  Behaviors in Dialog}, pages 15--18. University of Texas, September 2012.
\newblock ISBN not assigned.
\newblock null ; Conference date: 07-09-2012.

\bibitem[de~Kok et~al.(2013)de~Kok, Heylen, and Morency]{2013dekok}
Iwan de~Kok, Dirk Heylen, and Louis-Philippe Morency.
\newblock Speaker-adaptive multimodal prediction model for listener responses.
\newblock In \emph{Proceedings of the 15th ACM on International Conference on
  Multimodal Interaction}, ICMI '13, page 51–58, New York, NY, USA, 2013.
  Association for Computing Machinery.
\newblock ISBN 9781450321297.
\newblock \doi{10.1145/2522848.2522866}.
\newblock URL \url{https://doi.org/10.1145/2522848.2522866}.

\bibitem[Dickstein-Fischer et~al.(2018)Dickstein-Fischer, Crone-Todd, Chapman,
  Fathima, and Fischer]{dickstein2018socially}
Laurie~A Dickstein-Fischer, Darlene~E Crone-Todd, Ian~M Chapman, Ayesha~T
  Fathima, and Gregory~S Fischer.
\newblock Socially assistive robots: current status and future prospects for
  autism interventions.
\newblock \emph{Innovation and Entrepreneurship in Health}, 5:\penalty0 15--25,
  2018.

\bibitem[Gambino et~al.(2020)Gambino, Fox, and Ratan]{gambino2020building}
Andrew Gambino, Jesse Fox, and Rabindra~A Ratan.
\newblock Building a stronger casa: Extending the computers are social actors
  paradigm.
\newblock \emph{Human-Machine Communication}, 1:\penalty0 71--85, 2020.

\bibitem[Garau et~al.(2001)Garau, Slater, Bee, and Sasse]{2001garau}
Maia Garau, Mel Slater, Simon Bee, and Martina~Angela Sasse.
\newblock The impact of eye gaze on communication using humanoid avatars.
\newblock In \emph{Proceedings of the SIGCHI Conference on Human Factors in
  Computing Systems}, CHI '01, page 309–316, New York, NY, USA, 2001.
  Association for Computing Machinery.
\newblock ISBN 1581133278.
\newblock \doi{10.1145/365024.365121}.
\newblock URL \url{https://doi.org/10.1145/365024.365121}.

\bibitem[Gibson(1997)]{1997gibson}
Bob Gibson.
\newblock Taking the test: Using verbal report data in looking at the
  processing of cloze tasks.
\newblock \emph{Edinburgh Working Papers in Applied Linguistics}, 8:\penalty0
  54--62, 1997.

\bibitem[Gratch et~al.(2006)Gratch, Okhmatovskaia, Lamothe, Marsella, Morales,
  van~der Werf, and Morency]{Gratch2006}
Jonathan Gratch, Anna Okhmatovskaia, Francois Lamothe, Stacy Marsella, Mathieu
  Morales, R.~J. van~der Werf, and Louis-Philippe Morency.
\newblock \emph{Virtual Rapport}.
\newblock Springer Berlin Heidelberg, 2006.
\newblock ISBN 978-3-540-37594-4.
\newblock \doi{10.1007/11821830_2}.
\newblock URL \url{http://link.springer.com/10.1007/11821830_2}.

\bibitem[H.~Yngve(1970)]{1970yngve}
Victor H.~Yngve.
\newblock On getting a word in edgewise.
\newblock pages 567--577. Chicago Linguistic Society, 1970.

\bibitem[Hall and Watson(1970)]{1970hall}
Jay Hall and W.~H. Watson.
\newblock The effects of a normative intervention on group decision-making
  performance.
\newblock \emph{Human Relations}, 23\penalty0 (4):\penalty0 299--317, 1970.
\newblock \doi{10.1177/001872677002300404}.
\newblock URL \url{https://doi.org/10.1177/001872677002300404}.

\bibitem[Hart and Staveland(1988)]{1988nasatlx}
Sandra~G. Hart and Lowell~E. Staveland.
\newblock Development of nasa-tlx (task load index): Results of empirical and
  theoretical research.
\newblock In Peter~A. Hancock and Najmedin Meshkati, editors, \emph{Human
  Mental Workload}, volume~52 of \emph{Advances in Psychology}, pages 139--183.
  North-Holland, 1988.
\newblock \doi{https://doi.org/10.1016/S0166-4115(08)62386-9}.
\newblock URL
  \url{https://www.sciencedirect.com/science/article/pii/S0166411508623869}.

\bibitem[Huang et~al.(2019)Huang, Fukuda, and Nishida]{2019huang}
Hung-Hsuan Huang, Masato Fukuda, and Toyoaki Nishida.
\newblock Toward rnn based micro non-verbal behavior generation for virtual
  listener agents.
\newblock pages 53--63, 07 2019.
\newblock ISBN 978-3-030-21901-7.
\newblock \doi{10.1007/978-3-030-21902-4_5}.

\bibitem[Hunt and Thomas(2000)]{rubberduckbook}
Andrew Hunt and David Thomas.
\newblock \emph{The Pragmatic Programmer: From Journeyman to Master}.
\newblock Addison-Wesley Longman Publishing Co., Inc., USA, 2000.
\newblock ISBN 020161622X.

\bibitem[Hussain et~al.(2019)Hussain, Erzin, Sezgin, and Yemez]{2019hussain}
Nusrah Hussain, Engin Erzin, T.~Metin Sezgin, and Yucel Yemez.
\newblock Batch recurrent q-learning for backchannel generation towards
  engaging agents, 2019.
\newblock URL \url{https://arxiv.org/abs/1908.02037}.

\bibitem[Jain et~al.(2021)Jain, Leekha, Shah, and Shukla]{jain2021exploring}
Vidit Jain, Maitree Leekha, Rajiv~Ratn Shah, and Jainendra Shukla.
\newblock Exploring semi-supervised learning for predicting listener
  backchannels.
\newblock In \emph{Proceedings of the 2021 CHI Conference on Human Factors in
  Computing Systems}, pages 1--12, 2021.

\bibitem[John et~al.(1991)John, Donahue, and Kentle]{john1991big}
Oliver~P. John, Eileen~M. Donahue, and Robert~L. Kentle.
\newblock The big five inventory—versions 4a and 54, 1991.

\bibitem[Jung et~al.(2013)Jung, Lee, DePalma, Adalgeirsson, Hinds, and
  Breazeal]{2013jung}
Malte~F. Jung, Jin~Joo Lee, Nick DePalma, Sigurdur~O. Adalgeirsson, Pamela~J.
  Hinds, and Cynthia Breazeal.
\newblock Engaging robots: Easing complex human-robot teamwork using
  backchanneling.
\newblock In \emph{Proceedings of the 2013 Conference on Computer Supported
  Cooperative Work}, CSCW '13, page 1555–1566, New York, NY, USA, 2013.
  Association for Computing Machinery.
\newblock ISBN 9781450313315.
\newblock \doi{10.1145/2441776.2441954}.
\newblock URL \url{https://doi.org/10.1145/2441776.2441954}.

\bibitem[Kabaci{\'n}ska et~al.(2021)Kabaci{\'n}ska, Prescott, and
  Robillard]{kabacinska2021socially}
Katarzyna Kabaci{\'n}ska, Tony~J Prescott, and Julie~M Robillard.
\newblock Socially assistive robots as mental health interventions for
  children: a scoping review.
\newblock \emph{International Journal of Social Robotics}, 13\penalty0
  (5):\penalty0 919--935, 2021.

\bibitem[Kawahara et~al.(2016)Kawahara, Yamaguchi, Inoue, Takanashi, and
  Ward]{2016kawahara}
Tatsuya Kawahara, Takashi Yamaguchi, Koji Inoue, Katsuya Takanashi, and Nigel
  Ward.
\newblock Prediction and generation of backchannel form for attentive listening
  systems.
\newblock pages 2890--2894, 09 2016.
\newblock \doi{10.21437/Interspeech.2016-118}.

\bibitem[Kendon(1967)]{1967kendon}
Adam Kendon.
\newblock Some functions of gaze-direction in social interaction.
\newblock \emph{Acta Psychologica}, 26:\penalty0 22--63, 1967.
\newblock ISSN 0001-6918.
\newblock \doi{https://doi.org/10.1016/0001-6918(67)90005-4}.
\newblock URL
  \url{https://www.sciencedirect.com/science/article/pii/0001691867900054}.

\bibitem[Kennedy et~al.(2022)Kennedy, Baxter, Senft, Belpaeme, and
  Lemaignan]{2016kennedy}
J.~Kennedy, Paul Baxter, Emmanuel Senft, Tony Belpaeme, and S.~Lemaignan.
\newblock From characterising three years of hri to methodology and reporting
  recommendations.
\newblock volume 2016-April, pages 391--398, 2022.
\newblock ISBN 978-1-4673-8370-7.
\newblock \doi{10.1109/HRI.2016.7451777}.
\newblock URL \url{https://uwe-repository.worktribe.com/output/913325}.

\bibitem[Kontogiorgos et~al.(2020)Kontogiorgos, Pereira, Sahindal, van Waveren,
  and Gustafson]{2020kontogiorgos}
Dimosthenis Kontogiorgos, Andre Pereira, Boran Sahindal, Sanne van Waveren, and
  Joakim Gustafson.
\newblock Behavioural responses to robot conversational failures.
\newblock In \emph{Proceedings of the 2020 ACM/IEEE International Conference on
  Human-Robot Interaction}, HRI '20, page 53–62, New York, NY, USA, 2020.
  Association for Computing Machinery.
\newblock ISBN 9781450367462.
\newblock \doi{10.1145/3319502.3374782}.
\newblock URL \url{https://doi.org/10.1145/3319502.3374782}.

\bibitem[Ku and Ho(2014)]{2010Ku}
Kelly Ku and Irene Ho.
\newblock Metacognitive strategies that enhance critical thinking.
\newblock \emph{Metacognition and Learning}, 5:\penalty0 251--267, 05 2014.
\newblock \doi{10.1007/s11409-010-9060-6}.

\bibitem[Leighton(2009)]{2009leighton}
Jacqueline~P. Leighton.
\newblock Two types of think aloud interviews for educational measurement:
  Protocol and verbal analysis.
\newblock National Council on Measurement in Education, 2009.

\bibitem[Martinez-Martin et~al.(2020)Martinez-Martin, Escalona, and
  Cazorla]{martinez2020socially}
Ester Martinez-Martin, Felix Escalona, and Miguel Cazorla.
\newblock Socially assistive robots for older adults and people with autism: An
  overview.
\newblock \emph{Electronics}, 9\penalty0 (2):\penalty0 367, 2020.

\bibitem[Morency et~al.(2010)Morency, Kok, and Gratch]{2010morency}
Louis-Philippe Morency, Iwan Kok, and Jonathan Gratch.
\newblock A probabilistic multimodal approach for predicting listener
  backchannels.
\newblock \emph{Autonomous Agents and Multi-Agent Systems}, 20:\penalty0
  70--84, 01 2010.
\newblock \doi{10.1007/s10458-009-9092-y}.

\bibitem[Moshkina et~al.(2014)Moshkina, Trickett, and
  Trafton]{moshkina2014social}
Lilia Moshkina, Susan Trickett, and J~Gregory Trafton.
\newblock Social engagement in public places: a tale of one robot.
\newblock In \emph{Proceedings of the 2014 ACM/IEEE international conference on
  Human-robot interaction}, pages 382--389, 2014.

\bibitem[Mueller et~al.(2015)Mueller, Leuschner, Briem, Schmidt, Kilgour,
  Stueker, and Waibel]{2015mueller}
Markus Mueller, David Leuschner, Lars Briem, Maria Schmidt, Kevin Kilgour,
  Sebastian Stueker, and Alex Waibel.
\newblock Using neural networks for data-driven backchannel prediction: A
  survey on input features and training techniques.
\newblock In Masaaki Kurosu, editor, \emph{Human-Computer Interaction:
  Interaction Technologies}, pages 329--340, Cham, 2015. Springer International
  Publishing.
\newblock ISBN 978-3-319-20916-6.

\bibitem[Murray et~al.(2022)Murray, Walker, Nanavati, Alves-Oliveira, Filippov,
  Sauppe, Mutlu, and Cakmak]{2021murray}
Michael Murray, Nick Walker, Amal Nanavati, Patricia Alves-Oliveira, Nikita
  Filippov, Allison Sauppe, Bilge Mutlu, and Maya Cakmak.
\newblock Learning backchanneling behaviors for a social robot via data
  augmentation from human-human conversations.
\newblock In Aleksandra Faust, David Hsu, and Gerhard Neumann, editors,
  \emph{Proceedings of the 5th Conference on Robot Learning}, volume 164 of
  \emph{Proceedings of Machine Learning Research}, pages 513--525. PMLR, 08--11
  Nov 2022.
\newblock URL \url{https://proceedings.mlr.press/v164/murray22a.html}.

\bibitem[Nass and Moon(2000)]{2000nass}
Clifford Nass and Youngme Moon.
\newblock Machines and mindlessness: Social responses to computers.
\newblock \emph{Journal of Social Issues}, 56\penalty0 (1):\penalty0 81--103,
  2000.
\newblock \doi{https://doi.org/10.1111/0022-4537.00153}.
\newblock URL
  \url{https://spssi.onlinelibrary.wiley.com/doi/abs/10.1111/0022-4537.00153}.

\bibitem[O'Brien et~al.(2018)O'Brien, Cairns, and Hall]{2018obrien}
Heather O'Brien, Paul Cairns, and Mark Hall.
\newblock A practical approach to measuring user engagement with the refined
  user engagement scale (ues) and new ues short form.
\newblock \emph{International Journal of Human-Computer Studies}, 112, 04 2018.
\newblock \doi{10.1016/j.ijhcs.2018.01.004}.

\bibitem[Oertel et~al.(2013)Oertel, Cummins, Edlund, Wagner, and Campbell]{D64}
Catharine Oertel, Fred Cummins, Jens Edlund, Petra Wagner, and Nick Campbell.
\newblock D64: A corpus of richly recorded conversational interaction.
\newblock \emph{Journal on Multimodal User Interfaces}, 7\penalty0
  (1):\penalty0 19--28, 2013.

\bibitem[Oertel et~al.(2021)Oertel, Jonell, Kontogiorgos, Mora, Odobez, and
  Gustafson]{2021oertel}
Catharine Oertel, Patrik Jonell, Dimosthenis Kontogiorgos, Kenneth~Funes Mora,
  Jean-Marc Odobez, and Joakim Gustafson.
\newblock Towards an engagement-aware attentive artificial listener for
  multi-party interactions.
\newblock \emph{Frontiers in Robotics and AI}, 8:\penalty0 189, 2021.
\newblock ISSN 2296-9144.
\newblock \doi{10.3389/frobt.2021.555913}.
\newblock URL
  \url{https://www.frontiersin.org/article/10.3389/frobt.2021.555913}.

\bibitem[Okato et~al.(1996)Okato, Kato, Kamamoto, and Itahashi]{1996okato}
Y.~Okato, K.~Kato, M.~Kamamoto, and S.~Itahashi.
\newblock Insertion of interjectory response based on prosodic information.
\newblock In \emph{Proceedings of IVTTA '96. Workshop on Interactive Voice
  Technology for Telecommunications Applications}, pages 85--88, 1996.
\newblock \doi{10.1109/IVTTA.1996.552766}.

\bibitem[Petisca et~al.(2022)Petisca, Leite, Paiva, and Esteves]{2022petisca}
Sofia Petisca, Iolanda Leite, Ana Paiva, and Francisco Esteves.
\newblock Human dishonesty in the presence of a robot: The effects of situation
  awareness.
\newblock \emph{International Journal of Social Robotics}, 14:\penalty0 1--12,
  07 2022.
\newblock \doi{10.1007/s12369-022-00864-3}.

\bibitem[Poppe et~al.(2010)Poppe, Truong, Reidsma, and Heylen]{2010poppe}
Ronald Poppe, Khiet~P. Truong, Dennis Reidsma, and Dirk Heylen.
\newblock Backchannel strategies for artificial listeners.
\newblock In \emph{Proceedings of the 10th International Conference on
  Intelligent Virtual Agents}, IVA'10, page 146–158, Berlin, Heidelberg,
  2010. Springer-Verlag.
\newblock ISBN 3642158919.

\bibitem[Rabbitt et~al.(2015)Rabbitt, Kazdin, and
  Scassellati]{rabbitt2015integrating}
Sarah~M Rabbitt, Alan~E Kazdin, and Brian Scassellati.
\newblock Integrating socially assistive robotics into mental healthcare
  interventions: Applications and recommendations for expanded use.
\newblock \emph{Clinical psychology review}, 35:\penalty0 35--46, 2015.

\bibitem[Ramachandran et~al.(2018)Ramachandran, Huang, Gartland, and
  Scassellati]{2018ramachandran}
Aditi Ramachandran, Chien-Ming Huang, Edward Gartland, and Brian Scassellati.
\newblock Thinking aloud with a tutoring robot to enhance learning.
\newblock In \emph{Proceedings of the 2018 ACM/IEEE International Conference on
  Human-Robot Interaction}, HRI '18, page 59–68, New York, NY, USA, 2018.
  Association for Computing Machinery.
\newblock ISBN 9781450349536.
\newblock \doi{10.1145/3171221.3171250}.
\newblock URL \url{https://doi.org/10.1145/3171221.3171250}.

\bibitem[Rammstedt and John(2007)]{2007rammstedt}
Beatrice Rammstedt and Oliver~P. John.
\newblock Measuring personality in one minute or less: A 10-item short version
  of the big five inventory in english and german.
\newblock \emph{Journal of Research in Personality}, 41\penalty0 (1):\penalty0
  203--212, 2007.
\newblock ISSN 0092-6566.
\newblock \doi{https://doi.org/10.1016/j.jrp.2006.02.001}.
\newblock URL
  \url{https://www.sciencedirect.com/science/article/pii/S0092656606000195}.

\bibitem[Rudovic et~al.(2019)Rudovic, Zhang, Schuller, and Picard]{2019rudovic}
Ognjen Rudovic, Meiru Zhang, Bjorn Schuller, and Rosalind Picard.
\newblock Multi-modal active learning from human data: A deep reinforcement
  learning approach.
\newblock In \emph{2019 International Conference on Multimodal Interaction},
  ICMI '19, page 6–15, New York, NY, USA, 2019. Association for Computing
  Machinery.
\newblock ISBN 9781450368605.
\newblock \doi{10.1145/3340555.3353742}.
\newblock URL \url{https://doi.org/10.1145/3340555.3353742}.

\bibitem[Ruede et~al.(2019)Ruede, Müller, Stüker, and Waibel]{2019ruede}
Robin Ruede, Markus Müller, Sebastian Stüker, and Alex Waibel.
\newblock \emph{Yeah, Right, Uh-Huh: A Deep Learning Backchannel Predictor: 8th
  International Workshop on Spoken Dialog Systems}, pages 247--258.
\newblock 01 2019.
\newblock ISBN 978-3-319-92107-5.
\newblock \doi{10.1007/978-3-319-92108-2_25}.

\bibitem[Sacino et~al.(2022)Sacino, Cocchella, De~Vita, Bracco, Rea, Sciutti,
  and Andrighetto]{2022sacino}
Alessandra Sacino, Francesca Cocchella, Giulia De~Vita, Fabrizio Bracco,
  Francesco Rea, Alessandra Sciutti, and Luca Andrighetto.
\newblock Human- or object-like? cognitive anthropomorphism of humanoid robots.
\newblock \emph{PLOS ONE}, 17\penalty0 (7):\penalty0 1--19, 07 2022.
\newblock \doi{10.1371/journal.pone.0270787}.
\newblock URL \url{https://doi.org/10.1371/journal.pone.0270787}.

\bibitem[Scassellati et~al.(2012)Scassellati, Admoni, and
  Matari{\'c}]{scassellati2012robots}
Brian Scassellati, Henny Admoni, and Maja Matari{\'c}.
\newblock Robots for use in autism research.
\newblock \emph{Annual review of biomedical engineering}, 14:\penalty0
  275--294, 2012.

\bibitem[Serban et~al.(2015)Serban, Lowe, Henderson, Charlin, and
  Pineau]{2017corpora}
Iulian~Vlad Serban, Ryan Lowe, Peter Henderson, Laurent Charlin, and Joelle
  Pineau.
\newblock A survey of available corpora for building data-driven dialogue
  systems, 2015.
\newblock URL \url{https://arxiv.org/abs/1512.05742}.

\bibitem[Someren et~al.(1994)Someren, Barnard, and Sandberg]{1994vansomeren}
Maarten Someren, Yvonne Barnard, and Jacobijn Sandberg.
\newblock \emph{The Think Aloud Method - A Practical Guide to Modelling
  Cognitive Processes}.
\newblock 01 1994.

\bibitem[Stinessen(1985)]{1985stinessen}
Leif Stinessen.
\newblock The influence of verbalization on problem-solving.
\newblock \emph{Scandinavian Journal of Psychology}, 26\penalty0 (1):\penalty0
  342--347, 1985.
\newblock \doi{https://doi.org/10.1111/j.1467-9450.1985.tb01173.x}.
\newblock URL
  \url{https://onlinelibrary.wiley.com/doi/abs/10.1111/j.1467-9450.1985.tb01173.x}.

\bibitem[Truong et~al.(2010)Truong, Poppe, and Heylen]{2010truong}
{Khiet Phuong} Truong, {Ronald Walter} Poppe, and {Dirk K.J.} Heylen.
\newblock A rule-based backchannel prediction model using pitch and pause
  information.
\newblock In \emph{Proceedings of Interspeech 2010}, pages 3058--3061.
  International Speech Communication Association (ISCA), September 2010.
\newblock ISBN 1990-9772.
\newblock URL \url{http://www.interspeech2010.jpn.org/}.
\newblock null ; Conference date: 26-09-2010 Through 30-09-2010.

\bibitem[Ward and Tsukahara(2000)]{2000ward}
Nigel Ward and Wataru Tsukahara.
\newblock Prosodic features which cue back-channel responses in english and
  japanese.
\newblock \emph{Journal of Pragmatics}, 32\penalty0 (8):\penalty0 1177--1207,
  2000.
\newblock ISSN 0378-2166.
\newblock \doi{https://doi.org/10.1016/S0378-2166(99)00109-5}.
\newblock URL
  \url{https://www.sciencedirect.com/science/article/pii/S0378216699001095}.

\bibitem[Washburn et~al.(2020)Washburn, Adeleye, An, and Riek]{2020washburn}
Auriel Washburn, Akanimoh Adeleye, Thomas An, and Laurel~D. Riek.
\newblock Robot errors in proximate hri: How functionality framing affects
  perceived reliability and trust.
\newblock \emph{J. Hum.-Robot Interact.}, 9\penalty0 (3), may 2020.
\newblock \doi{10.1145/3380783}.
\newblock URL \url{https://doi.org/10.1145/3380783}.

\bibitem[Weizenbaum(1966)]{1966weizenbaum}
Joseph Weizenbaum.
\newblock Eliza—a computer program for the study of natural language
  communication between man and machine.
\newblock \emph{Commun. ACM}, 9\penalty0 (1):\penalty0 36–45, jan 1966.
\newblock ISSN 0001-0782.
\newblock \doi{10.1145/365153.365168}.
\newblock URL \url{https://doi.org/10.1145/365153.365168}.

\bibitem[Wijnen et~al.(2019)Wijnen, Davison, Reidsma, Meij, Charisi, and
  Evers]{2019wijnen}
Frances~M. Wijnen, Daniel~P. Davison, Dennis Reidsma, Jan Van~Der Meij, Vicky
  Charisi, and Vanessa Evers.
\newblock Now we’re talking: Learning by explaining your reasoning to a
  social robot.
\newblock \emph{J. Hum.-Robot Interact.}, 9\penalty0 (1), dec 2019.
\newblock \doi{10.1145/3345508}.
\newblock URL \url{https://doi.org/10.1145/3345508}.

\end{thebibliography}

\pagebreak
\onecolumngrid
\clearpage

\begin{center}
    \textbf{\large[Supplementary Material] \\
Robot Duck Debugging: Can Attentive Listening Improve Problem Solving?}

\end{center}

In order to assess how the user perceived the impact of thinking-aloud, we formulated 7 questions with a 5-point Likert scale (\textit{strongly disagree} to \textit{strongly agree}). The questions were asked in the post-experiment questionnaire to participants in all conditions (\textit{RDuck, NaïveL, DataL}). The questions are as follows:

\begin{enumerate}
    \item Thinking aloud has helped me complete the tasks.	
    \item Thinking aloud has not impacted my performance in these tasks.	
    \item Thinking aloud has improved my performance in these tasks.	
    \item Thinking aloud has improved my ability to think critically about my decisions during these tasks.	
    \item Thinking aloud has not influenced my thought process while solving these tasks.	\item Thinking aloud has made me more confident about my decisions during these tasks.	\item Thinking aloud has distracted me from the tasks.
\end{enumerate}

Questions $Q2$ and $Q5$ were removed from the final \textbf{Impact of Thinking Aloud} index, as their valence is ambiguous. The final index is given by $(Q1 + Q3 + Q4 + Q6 + [6-Q7]) / 5$ .

A Cronbach's Alpha of $\alpha=0.84$ indicated good reliability.

\end{document}